\documentclass[letterpaper, 10 pt, conference]{ieeeconf}  

\IEEEoverridecommandlockouts                              

\usepackage{graphics} 
\usepackage{graphicx}
\usepackage{subcaption}
\usepackage{amsmath} 
\usepackage{multirow}
\usepackage{booktabs}
\usepackage{cite}

\title{\LARGE \bf
A Multi-Behavior Planning Framework for Robot Guide
}

\author{Muhan Hou$^{1}$, Zonghao Mu$^{1}$, Jing Li$^{2}$,  Qizhi Yu$^{1}$, Jason Gu$^{1}$
\thanks{$^{1}$Muhan Hou, Zonghao Mu, Qizhi Yu and Jason Gu are with Zhejiang Lab, Hangzhou, China
        {\tt\small \{houmh,muzh,yuqz,jgu\}@zhejianglab.com}}%
\thanks{$^{2}$Jing Li is with School of Computer Science and Engineering, Tianjin University of Technology, Tianjin, China}%
}

\begin{document}

\maketitle
\thispagestyle{empty}
\pagestyle{empty}

\begin{abstract}

The guiding task of a mobile robot requires not only human-aware navigation, but also appropriate yet timely interaction for active instruction. State-of-the-art tour-guide models limit their socially-aware consideration to adapting to users' motion, ignoring the interactive behavior planning to fulfill the communicative demands. We propose a multi-behavior planning framework based on Monte Carlo Tree Search to better assist users to understand confusing scene contexts, select proper paths and timely arrive at the destination. To provide proactive guidance, we construct a sampling-based probability model of human motion to consider the interrelated effects between robots and humans. We validate our method both in simulation and real-world experiments along with performance comparison with state-of-the-art models.

\end{abstract}

\section{Introduction}

Guiding human through an environment (e.g., museums \cite{IEEEexample:1}, airports \cite{IEEEexample:2} and shopping malls \cite{IEEEexample:3}) with a mobile service robot has long been an active topic in human-robot interaction (HRI). The goal of guiding is to lead users to the right destination along a proper route and within a reasonable amount of time. Therefore, the guide robots should treat human partners as objectives to serve with socially compliant consideration rather than obstacles to avoid. Navigation algorithms that are adaptive to the users have been pursued in previous work \cite{IEEEexample:2, IEEEexample:4, IEEEexample:5, IEEEexample:6}. However, due to the lack of active instructions such as human-like pointing, it is still prone for human partners to get confused about robots' intention.

Humanoid mobile robots, equipped with rotatable heads, movable arms, and speakers, are increasingly used for tour-guide services. These robots are capable of performing vivid guiding expressions, which can help users better understand robots' guidance intention. However, dealing with multiple behaviors makes the robot planning task more complicated. How to optimally select human-like behaviors across multiple modalities yet remains an open question. Robot guiding with unnecessary active instructions causes inefficiency and may fail the guide task due to overtime arrival. Moreover, social discomfort may be induced by excessive active instruction. Rule-based behavior planning \cite{IEEEexample:7} is only suitable for the applications without the need of user adaptation. Recently, Dugas et al. \cite{IEEEexample:8} introduced a multi-behavior planner for navigation among humans based on Monte Carlo Tree Search (MCTS). Neither of them is sufficient for the guiding task.
\begin{figure}[thpb]
      \centering
      \includegraphics[width=2.5in]{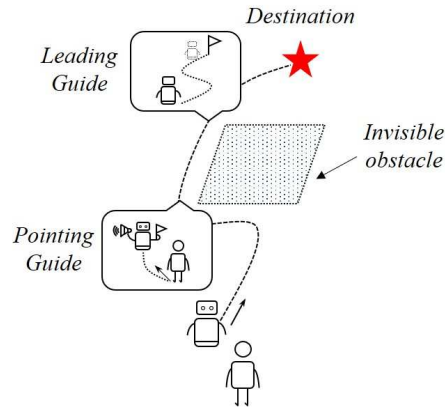}
      \caption{Demonstration of the multi-behavior planning. The area in shadow represents the space to avoid without obvious notification. The robot would optimally organize different guiding behaviors to help the user understand confusing scene contexts and take necessary detours. }
      \label{figurel}
\end{figure}

For the guiding purpose, behavior planning across various modalities must balance between efficiency and legibility. To be clear, legibility \cite{IEEEexample:9} refers to whether the intention of robot behavior can be fully understood by humans, and efficiency refers to whether human can be guided to the destination in time. Specifically, two aspects of problems are of significant challenges: 1) how to predict human motion considering the influence of robot guiding behaviors; and 2) how to select optimal behavior sequences to assist users to pick a proper path and arrive at the destination in time.

Inspired by Dugas et al. \cite{IEEEexample:8}, this paper attempts to optimize the behavior selection policy by a repetitive prediction-selection procedure. At every time step, we predict long-term human motions in response to consecutive robot guiding behaviors. The optimal action sequence is selected based on MCTS to help humans better understand robot active instruction and decide the proper timing and position to switch between different behaviors, as shown in Fig. 1.

In summary, our contributions are as follow:
\begin{itemize}
\item We propose a multi-behavior planning framework for providing proactive guidance.

\item We design a sampling-based prediction model of human motion that explicitly depicts potential impact from legibility of various robot guiding behaviors.

\item We validate our method in both simulation and real-world environments.
\end{itemize}

\section{Related Work}

\subsection{Robot Guide}

Guiding users by a robot through an environment is a challenging topic in HRI research. Robot adaptivity to human motion has been improved in various aspects, including path, velocity and moving pattern. To enhance the adaptivity in path selection, Pandey and Alami \cite{IEEEexample:4} planned the trajectory via a multi-variant Gaussian model and switched between different guiding modes for corresponding human behaviors. Nakazawa et al. \cite{IEEEexample:5} and Zhang et al.\cite{IEEEexample:6} tackled the path planning problem based on artificial potential field, realizing more natural mode transitions with one unified framework. To improve robot adaptivity in migration velocity, a speed adjustment strategy \cite{IEEEexample:2} was proposed based on hierarchical Mixed Observability Markov Decision Process (MOMDP) for high-level decision making. To be more adaptive to human comfort in moving pattern, Shiomi et al. \cite{IEEEexample:10} designed a tour-guide with a walking-backward fashion, which much facilitates interaction behaviors and attracts more bystanders to overhear the speech. However, navigation alone could hardly distinguish ambiguous scene contexts for human partners. Its deficiency in intention legibility would much possibly lead to human confusion about path selecting.

\subsection{Multi-Behavior Planning}

To make robot intention more understandable, more interactive behaviors are required. This will expand the dimension of robot behaviors beyond navigation, calling for proper planning and decision-making policy. Previous works tried to organize behaviors across multiple modalities with either short-term consideration or long-term trade-off. From the aspect of short-term criterion, simple vocal request was added in \cite{IEEEexample:2} and \cite{IEEEexample:4} to confirm inquiry or termination whenever human states are abnormal. Kanda et al. \cite{IEEEexample:3} transitioned among behaviors of guiding, socializing, and advertising via the Wizard of Oz technique, i.e., manually operating robots by humans behind the scenes. These case-by-case methods tried to maximize immediate interest and could barely be applied when proactive actions are expected. Also, modal differences in legibility and efficiency were not fully considered solely via intuitive switching among various behaviors. When it turns to long-term balance, Nishimura and Yonetani \cite{IEEEexample:11} constructed a deep Reinforcement Learning (RL) framework to decide the timing of addressing path-clear request as a substitute for bypass navigation. Dugas et al. \cite{IEEEexample:8} selected among three customized navigation behaviors using MCTS to navigate through congested environments. However, the interplay between humans and robots was not sufficiently evaluated. The influence from diverse robot behaviors on human states was simply taken as deterministic without further examination. It may become invalid when tasks demand joint effort, as the case in tour-guide planning.

\subsection{Human Motion Prediction}

In order to provide proactive actions, predicting future human motion is indispensable. When robots are coexistent with humans, existing methods tried to foresee human movement either via independent planning, or integrating them with robots as a joint group. Modeling each human as an independent agent, Bruckschen et al. \cite{IEEEexample:12} predicted human navigation goal via a Bayesian-inference based interaction model. Rudenko et al. \cite{IEEEexample:13} fit human motion into Markov Decision Process (MDP) and a revised social force model, predicting human goals and trajectories based on probabilistic sampling. Fahad et al. \cite{IEEEexample:14} directly extracted human motion policy from pedestrian datasets and solved the prediction problem in a model-free way. By contrast, some other methods paired humans and robots as an integral and corporately predicted future motion pattern. They constructed the joint state space shared by humans and robots and matched possible actions with various configurations by means of Graph Convolutional Network (GCN) \cite{IEEEexample:15}, Partially Observable Markov Decision Process (POMDP) \cite{IEEEexample:16} and the game theory \cite{IEEEexample:17}. However, the influence of robot behaviors was seldomly discussed explicitly. Specifically, characteristic differences of various modalities, such as intention legibility and time efficiency, were not clearly identified, leaving a blank for further investigation.

\section{Method}

We propose a multi-behavior planning framework to perform proper guiding behaviors in appropriate time so that human partners can arrive at the right destination along a proper route and within a reasonable amount of time. With explicit consideration of the influence of robot guidance, human motion is predicted to enable the robot with proactivity to envision long-term human movement.



In this work, we select two most-often used guiding behaviors, i.e., \textit{leading guide} and \textit{pointing guide}, to demonstrate our multi-behavior framework. The leading guide refers to navigation movement without any other interaction. The pointing guide includes speech and gestures for proactive expression of intentions. Note that our framework can be generalized by combining more types of behaviors

\subsection{Guiding Behavior Planning Framework}

Given the state of human and robot at the current moment, decisions need to be made regarding where the robot should move and how it should behave along the way. We pair the human-robot state into a node and conduct a tree-shaped search for an optimal action-position sequence.

By making sufficient attempts with different behavior sequences and envisioning their outcomes, it seems more possible to find the one with optimal guide performance. However, the computational cost would also surge dramatically without fully leveraging previous experience. Conventional tree search methods, like Breadth-first search or Depth-first search, are either inefficient in exploitation or insufficient in exploration. Instead, we formulate our problem via MCTS and produce a sampling-based directional search. It can better solve exploitation-exploration dilemma and find the optimal policy to switch between different guiding behaviors.

\subsubsection{Formulation}

Starting from the current human-robot state, we set this state pair as the root node and build the tree thereafter. Following the pipeline of MCTS,  each trial of search process includes four consecutive steps: selection, expansion, simulation, and update. Trials will be repeated until reaching certain time limit and produce an optimal policy, as shown in Fig. 2.

The selection step is designed to determine the direction along which the search process will further explore.  It will be executed on each depth and transit to the child node with the highest score until a leaf node is reached. To be specific, we define the score value of each node, as:
\begin{equation}
        V_s(s_i)=-\overline{C}_{final,i} + c\cdot \sqrt{\frac{ln(N)}{n_i}},
\end{equation}
where $s_i$ represents the configuration of human-robot state pair, $\overline{C}_{final,i}$ is the average of the cumulative final weighted cost of the node to be evaluated, $c$ is a constant for explore-exploit balance, $N$ is the visit times for the parent node and $n_i$ is the visit times for the selected node. It is obvious that the priority would be reserved for nodes that have never been explored before. In that case, we would always select the first one among children nodes without losing generality.

The expansion step is used to supplement more possible situations along the direction selected before. It will follow and expand the tree with children nodes sampled from certain specific areas. Among the information contained in each child node, the human state is obtained from our motion prediction algorithm, which will be explained in subsection B. The prediction outcomes vary according to diverse selections of the next robot position and the guiding behavior along the way. The impact from legibility differences of robot actions will be reflected in the prediction process.

The simulation step is applied to envision the outcome of the explored node sequence. A random rollout policy will be followed to obtain future human-robot states until a success is acclaimed or reaching terminal conditions.
\begin{figure}[thpb]
      \centering
      \includegraphics[width=2.5in]{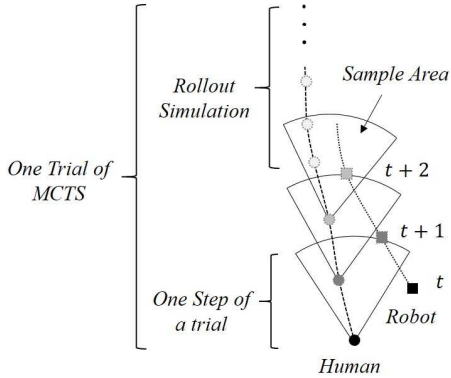}
      \caption{MCTS-based behavior planning process. The solid black circle and solid black square represent the real position of human and robot for the current moment respectively. The more distant it is in time, the lighter the color shows.  Each human-robot pair at the same time depth constitutes a node of the search tree. A rollout simulation will follow on after arriving at a leaf node and terminate as mission completed or at max depth.}
      \label{figure2}
\end{figure}

Finally, the update step will follow to backwardly feedback the consequent cost. The cumulative cost of the explored sequence will hereby be evaluated and get updated for each node along the way.

\subsubsection{Implementation}

\textit{Behavior-oriented Sample Area} The sample area used in the expansion step to supplement possible future situations are defined as behavior-oriented sample area. It is a fan-shape area centered at the given human position, as shown in Fig. 3. Furthermore, the resultant sampling zone is symmetric about the line joining the human position and the robot guiding destination with the angle range of $\theta_s \in [0, \pi]$.

The sampling would only be implemented along the arcs with radius valued as appropriate social distances and sampled at a small interval of $\Delta\theta \in [0, \pi]$. Considering the difference between migrating velocities of different robot guiding behaviors, radiuses vary accordingly for different behavior selections.

\textit{Cost Function} In order to guide humans to reach the destination along a proper path, within the time limit, and avoiding certain areas (e.g., invisible obstacles or hazards), three costs are defined: \textit{distance cost}, \textit{time cost} and \textit{affordance cost}.

Given a target human path:
$$
T_{target}=\left\{ p_0^{target}, p_1^{target}, p_2^{target}, ..., p_{l_{target}-1}^{target} \right\}
$$
where $l_{target}$ is the length of the $T_{target}$, and a real human path:
$$T_{real}=\left\{p_0^{real}, p_1^{real}, p_2^{real}, ..., p_{l_{real}-1}^{real} \right\}
$$
where $l_{real}$ is the length of the $T_{real}$, the \textit{distance cost} for the current time step $t$ can be defined as:
\begin{equation}
          \begin{aligned}
          C_{dist}^t = \sum_{i=0}^{N-1}\left\{\zeta_d(i)\cdot\|p_{i+l_{target}-n_g}^{target}-p_i^{real}\|+ \right.\\
          \left.\left[1-\zeta_d(i)\right]\cdot\|p_{l_{target}-1}^{target}-p_i^{real}\|\right\},
          \end{aligned}
\end{equation}
where $n_g$ is the minimum search depth required to reach the goal, representing the minimum future steps the human is expected to take following the target human path, $N=max(l_target,n_g)$ represents the number of check points required for calculating the distance cost. $\zeta_d$ is the flag parameter for proper index switching between $T_{target}$ and $T_{real}$:
\begin{equation}
        \zeta_d(i)=\left\{
        \begin{aligned}
            &1,\  if \ i+l_{target}-n_g < l_{target} \\
            &0,\  else
        \end{aligned}
        \right.,
\end{equation}
It should be noted that the adjacent points, both in $T_{target}$ and $T_{real}$, are at the same time interval of $t_{per-step}$, instead of equal geography distances. For every time step, the MCTS would feedback a set of candidate policies after considerable episodes of exploitation and exploration, and $C_{dist}^t$ would be calculated along with the time cost at the end of each iteration.

In addition to the distance cost, time cost plays a role in the selection of optimal behavior sequence as well. However, unlike the distance cost, which would be existent and examined at each check point, time cost would only be applicable when the depth of the current human-robot state node is beyond the value of nodes-to-go. Therefore, the \textit{time cost} at time step t can be expressed as:
\begin{equation}
        C_{time}^t=\sum_{i=0}^{l_{real}-1}\zeta_t(i)t_{per-step},
\end{equation}
where $\zeta_t$ is the flag parameter for applicability check of the time cost for each point in $T_{real}$ and defined as:
\begin{equation}
        \zeta_t(i)=\left\{
        \begin{aligned}
            &1, \ if \ i > n_g \\
            &0, \ if \ i \leq n_g
        \end{aligned}
        \right.,
\end{equation}

In order to assist human to select proper paths and take necessary detours against certain areas, an \textit{affordance cost} is constructed as:
\begin{equation}
        C_{aff}^t=\sum_{i=0}^{l_{real}-1}\zeta_{aff}(i)\cdot C_0,
\end{equation}
where $\zeta_{aff}$ is the flag of applying the affordance cost:
\begin{equation}
        \zeta_{aff}(i)=\left\{
        \begin{aligned}
            &1, \ if \ T_{real}(i) \in S_{aff} \\
            &0, \ otherwise
        \end{aligned}
        \right.,
\end{equation}
Whenever the resultant human position is within spaces to avoid, i.e., $S_{aff}$, a constant affordance cost $C_0$ will be superimposed to encourage proactive and proper selection for both the timing and the position to perform different guiding behaviors.

Taking into account the significance trade-off between the distance cost and the time cost, the \textit{final weighted cost} can be simply formulated as:
\begin{equation}
        C_{final}^t=w_d \cdot \frac{C_{dist}^t}{k_d} + w_t \cdot C_{time}^t + w_{aff} \cdot C_{aff}^t,
\end{equation}
where $w_d$, $w_t$ and $w_{aff}$ are the associated weights of \textit{distance cost}, \textit{time cost} and \textit{affordance cost} respectively. $k_d$ is the efficient for the normalization of distance cost.
\begin{figure}[thpb]
      \centering
      \includegraphics[width=2.5in]{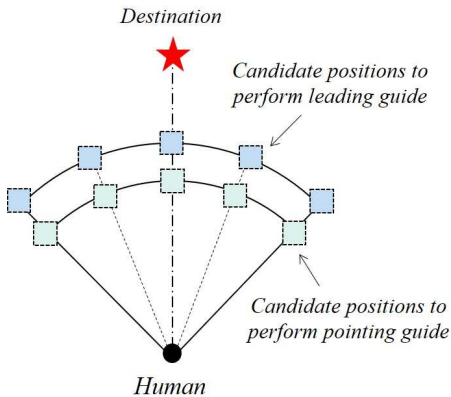}
      \caption{Behavior-oriented sample area}
      \label{figure3}
\end{figure}
\subsection{Human Motion Prediction}

Human motion can be influenced by robot guidance during tour-guide. We explicitly reflect the legibility impact of different robot behaviors with a \textit{legibility gain} and predict human motion via a sampling-based algorithm with the probability distribution characterized by MDP and a revised Social Force Model \cite{IEEEexample:18}. Furthermore, we define the human subjective motion as \textit{active motion} and the robot-influenced movement as \textit{passive motion}.

Although Rudenko et al. \cite{IEEEexample:13} realized the long-term prediction of pedestrian movement using MDP and social force model, they did not take into account the characteristic variation (e.g., legibility) of multiple modalities and its influence on future human motion. Additionally, their linear addition of MDP-based sample results and the social force assumes the potential interplay between moving agents as deterministic. It may not be true in certain uncooperative cases, like the agent insisting walking away although the robot is vocally requesting for attention and physically pointing towards the destination.

\subsubsection{Active Motion}

For each candidate goal $g \in G$, we formalize the human active motion towards it as an independent MDP model.

Given the gridmap of the environment, each single grid would be a state $(x,y)$ of the MDP model for a given goal $g \in G$.  At every state, the human agent is able to move at certain velocity $v \in [0, v_{max}]$ towards certain direction $\theta \in [0, 2\pi]$. We pair the two basic motion variables as an action $a = (v, \theta) \in A$ the agent could take at state $s \in S$.

The reward for each state can be divided into two separate parts: \textit{occupation cost} and \textit{action effort cost}. For the \textit{occupation cost}, whenever a grid cell is occupied, the cost $C_0 = 10$, otherwise it would be $0$ for free states. For the \textit{action effort cost}, it is positively related to the Euclidean distance that the selected action would cover. However, it is worth noting that this only holds true for non-goal states. For the goal state, the action effort cost would be $0$ for every possible action.

Therefore, we would obtain our reward function for the given goal $g \in G$ as follows:
\begin{equation}
        R_g(s,a)=\left\{
        \begin{aligned}
            &-\alpha C_0(S')-(1-\alpha)C_e(s,a), s \neq g  \\
            & 0, s=g
        \end{aligned}
        \right.,
\end{equation}
where $s'$ is the consequent state translated after taking action $a$ at state $s$, and $C_o(s')$ is the corresponding \textit{occupation cost} for the resulting state. Similarly, $C_e(s,a)$ is the action effort cost after the agent takes action at state $s$ and $\alpha$ is the controlling coefficient to balance the relative importance between the occupation cost and the action effort cost.

Considering the reality of human motion, we construct the transition as deterministic after taking action $a = (v, \theta)$ at state $s = (x, y)$ and translating to a new state $s' = (x', y')$, which means $T(s,a,s') = 1$. Furthermore, according to the definition of states and action, the resulting new state $s'$ satisfies:
\begin{equation}
        \left \{
        \begin{aligned}
            x' = x + v \cdot cos(\theta) \\
            y' = y + v \cdot sin(\theta)
        \end{aligned}
        \right.,
\end{equation}

For each given goal $g \in G$, we solve the MDP problem via value iteration and would obtain the resulting optimal value function $V_g^* (s)$ with the optimal policy $\pi_g^*$ as follows:
\begin{equation}
        \left \{
        \begin{aligned}
             Q_g^*(s,a) &= R_g(s,a) + \gamma \sum_{s'}p(s'|s,a)V_g^*(s') \\
             V_g^*(s) &= \max_a Q_g^*(s,a) \\
             \pi_g^*(s) &= \mathop{\arg\!\max}_{a}Q_g^*(s,a)
        \end{aligned}
        \right.,
\end{equation}
The resulting optimal policy represents the independent active motion of the human agent when the robot is absent from the scene. However, the prediction of human movement might be influenced and different when the robot comes in and will be reflected in the \textit{passive motion}.

\subsubsection{Passive Motion}

Compared with the active motion, the passive motion embodies the potential impact from the robot behaviors on future human movement. This is expressed via Social Force Model [18] and would play a role in the probability distribution of future human actions.

Similar to Rudenko et al. \cite{IEEEexample:12}, we construct the social force exerted by the robot on the human as:
\begin{equation}
        f_{h,r} = e^{\frac{d_{h,r} - d_{social}}{k_n}}\left\{\lambda + (1-\lambda)\frac{1+cos(\varphi_{h,r})}{2}\right\}\cdot\vec{n}_{h, r},
\end{equation}
where $f_{h, r}$ is the social force between the human and the robot, $d_{h, r}$ is the distance between these two agents, $d_{social}$d is the human social distance, and $k_n$ is used to normalize the magnitude of the resulting force. In addition to the relative distance, the relative direction of the human and the robot would also play a part in the force magnitude. This is inherited by $\varphi_{h, r}$, which represents the angle between the direction of the observed human velocity and the direction from the human to the robot. The last variable $\vec{n}_{h, r}$ is a unit vector pointing from the human to the robot, inferring the direction information of the consequent social force.

\subsubsection{Sampling-based Human Motion Prediction}

For every time instant, two-fold prediction is required: i) the current goal of human motion, ii) the next position the human may arrive at towards the selected goal.

\textit{Goal Prediction} For each goal $g \in G$, the optimal state values $V_g^*(s)$ for each state $s$ in the grid map can be solved following the MDP setup mentioned before. Assuming the past trajectory record $T_{human}^t = (s_0^t, s_1^t, s_2^t, ..., s_l^t)$ of fixed length $l$ is maintained and updated at every time step, the latest state $s_l^t$ and the initial past state $s_0^t$ can then be extracted from the sequence. Therefore, we construct the possibility of the goal $g \in G$ selected by human as the final destination for time step $t$ as:
$$
p^t(g) \propto
$$
\begin{equation}
        \left \{
        \begin{aligned}
            exp & \left(  \frac{\beta_g \left[V_g^*(s_l^t)-V_g^*(s_0^t)\right]}{\xi\left(1+\frac{f_{h,r}^i}{\sum_k f_{h,r}^k}\right)}\right), \ if \  g=g_{robot} \\
            \\
            exp & \left(  \beta_g \left[V_g^*(s_l^t)-V_g^*(s_0^t)\right]\right), \ if \  g \neq g_{robot}
        \end{aligned}
        \right.,
\end{equation}
where $\beta_g$ is the factor for goal probability normalization and $\xi > 1$ is the \textit{legibility gain} for the robot guiding behaviors. When the goal $g$ is the same as the destination $g_{robot}$, different robot behaviors would increase the goal probability by different extents. We explicitly depict such variation via the \textit{legibility gain} $\xi$. Additionally, the relative position of the robot would also have influence on the probability of $g_{robot}$. This is reflected by the social force $f_{h, r}^i$ and its weight among all the position options for the robot.

\textit{Position Prediction} Similar to the goal prediction, the possibilities of choosing candidate actions at the current state is also related to both the state values in MDP and the social force exerted by the robot.

Assuming the human is at state s, the previous optimal action-value function under the MDP setup for the goal $g \in G$ is then revised as:
\begin{equation}
        \widetilde{Q}_g^*(s,a) = w \cdot R_g(s,a) + \gamma \cdot V_g^*(s'),
\end{equation}
where $\widetilde{Q}_g^*(s,a)$ is the revised sub-optimal action-value function for the $g \in G$ after the human agent takes action $a$ from state $s$. The parameter $w$ is the control factor to encourage the human agent to take less optimal actions, which could often be the case when the human is walking around in an open environment without fixed and a-priori interest of some certain area.

Thereby, the possibility of selecting action $a$ at state $s$ under goal setup $g \in G$ can be constructed as:
$$
        a \sim \widetilde{\pi}_g(s) \ with \  prob \propto
$$
\begin{equation}
        \left \{
        \begin{aligned}
            exp & \left(\frac{\beta_a \left[\widetilde{Q}_g^*(s,a)-V_g^*(s')\right]}{\xi\left(1+\frac{f_{h,r}^i}{\sum_k f_{h,r}^k}\right)}\right), \ if \ s' \in S_f^t \\
            \\
            exp & \left(\beta_a \left[\widetilde{Q}_g^*(s,a)-V_g^*(s')\right]\right), \ if \ s' \notin S_f^t
        \end{aligned}
        \right.,
\end{equation}
where $\beta_a$ is the factor for action probability normalization. $\xi$ is the same \textit{legibility gain} as that used for goal probability, reflecting various attraction effects from different guiding behaviors.
\begin{figure}[thpb]
      \centering
      \includegraphics[width=2.5in]{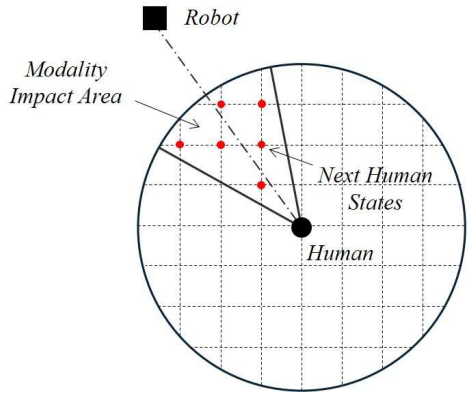}
      \caption{The area of modality impact. The upper-left fan-shaped area indicates the domain where future human motion will be impacted by robot guiding behaviors. Next states falling into this region will have higher possibility to be selected as a result of the robot impact.}
      \label{figure4}
\end{figure}
It is worth noting that the modality impact would only apply to the new states that fall into the influence range of the robot behavior at current time step, represented by $S_f^t$. As shown in Fig. 4, the modality impact area is defined as a fan-shaped region. This impact area is symmetric about the line connecting the human and the selected robot position and covers a constant total angle range of $\theta_m \in [0, \pi]$. Therefore, when the robot is performing guiding behaviors, human future positions within the area of modality impact would have relative higher possibilities to be selected. When it is beyond the reach of modality impact, future human motion would only depend on subjective willingness, i.e., the active motion.

\textit{Sampling-based Prediction Algorithm} With the pre-solved optimal state values $V_g^*$ via (3) for each MDP setup of candidate goals $g \in G$, the probability distribution of the goals can be obtained via (5). For each sample, we randomly choose a goal conforming to its possibility and follow it with an action selection against the probability calculated by (7). The resulting position corresponding to the new state $s'$ would then be marked in the layer $L$, shared by all the sample iterations, and complete one sample process for prediction. After $K$ times repeated sampling, the layer $L$ would store all the previous prediction results of next human positions. We normalize the result into the form of probability and do a final sample to select a position as the final prediction result at time instant $t$.

\section{Experiments and Results}

In order to validate our multi-modal behavior planning framework, we tested the method in the environment of an exhibition hall. Experiments were conducted in both simulation (Fig. 5) and real-world environments (Fig. 6) with three different scene settings. In each scenario, there are four different exhibition boards representing the candidate goals for the human agent.  The guide robot's task is to lead a visitor toward one of the exhibition board with optimal sequence of guiding behaviors pursuing legible and efficient user experiences.

For further quantitative comparison, three metrics were introduced to evaluate the guide-planning performance: 1) overall guide success rate, 2) context-ambiguity disturbed time ratio, and 3) socially discomfort time ratio.

The overall guide success rate was defined as the ratio between the trials where the robot leads the human partner to successfully arrive at the predefined destination and the total trials of the specific scene. A success would be claimed if the human partner was guided to arrive at the goal within a certain time limit.

The context-ambiguity disturbed time ratio was defined as the proportion of the time human staying in the inconspicuous affordance spaces versus the total time of a single trial. This would indicate how well a guide-behavior planner performs to distinguish the context ambiguity for the human agent.

The socially discomfort time ratio was defined as the ratio between the time human might feel uncomfortable due to proxemics consideration and that of the whole trial. The socially discomfort time would be added up whenever the distance between the human agent and the robot falls beneath certain distance threshold. Similar to \cite{IEEEexample:10}, two distance thresholds were defined: personal distance $d_p$ and intimate distance $d_i$, which are valued as $1.2m$ and $0.45m$ respectively.

\subsection{Simulation Experiments}

We tested our method in the simulation environment with three different space configurations (Fig.5). Parameter optimization was performed in a hand-tuning fashion with the following results: ($c$, $\theta_s$, $\Delta\theta$, $l_{target}$, $l_{real}$, $C_0$, $w_d$, $k_d$, $w_t$, $w_{aff}$) = ($1$, $\frac{\pi}{3}$, $\frac{\pi}{10}$, $2$, $2$, $10$, $1$, $100$, $1$, $1$) for the behavior planning framework and ($d_{social}$, $\theta_m$, $\lambda$, $l$, $\beta_g$, $\beta_a$, $w$, $\gamma$) = ($2$, $\frac{\pi}{3}$, $1$, $2$, $0.5$, $0.5$, $1$, $0.9$) for the human motion prediction. We set the legibility gain $\xi$ equal to $2$ and $4$ for leading guide and pointing guide respectively, reflecting different influence of various robot guiding behaviors on human motion.
\begin{figure}[thpb]
      \centering
	   \begin{subfigure}[b]{0.9\linewidth}
    	  \includegraphics[width=\linewidth]{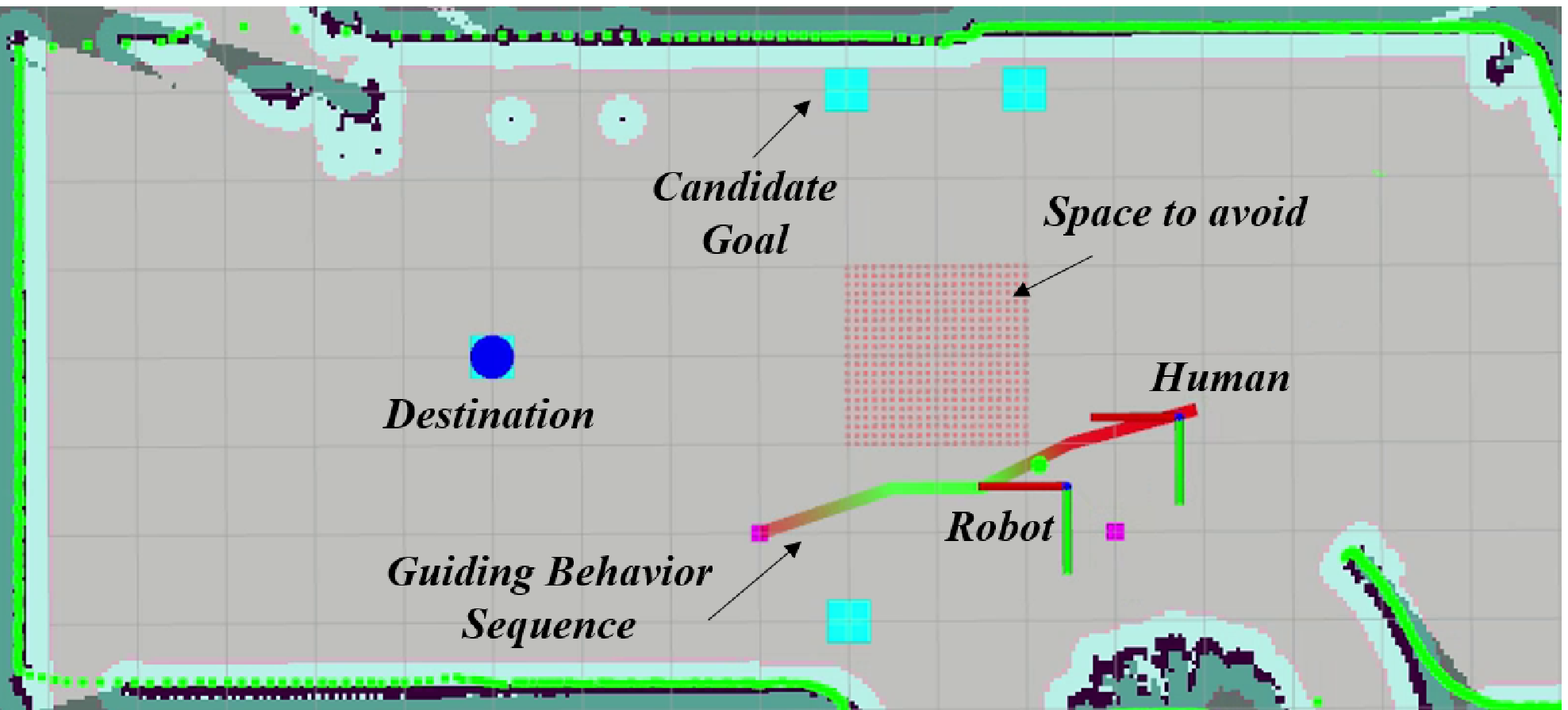}
         \caption{Scene 1}
 	   \end{subfigure}
       \begin{subfigure}[b]{0.9\linewidth}
         \includegraphics[width=\linewidth]{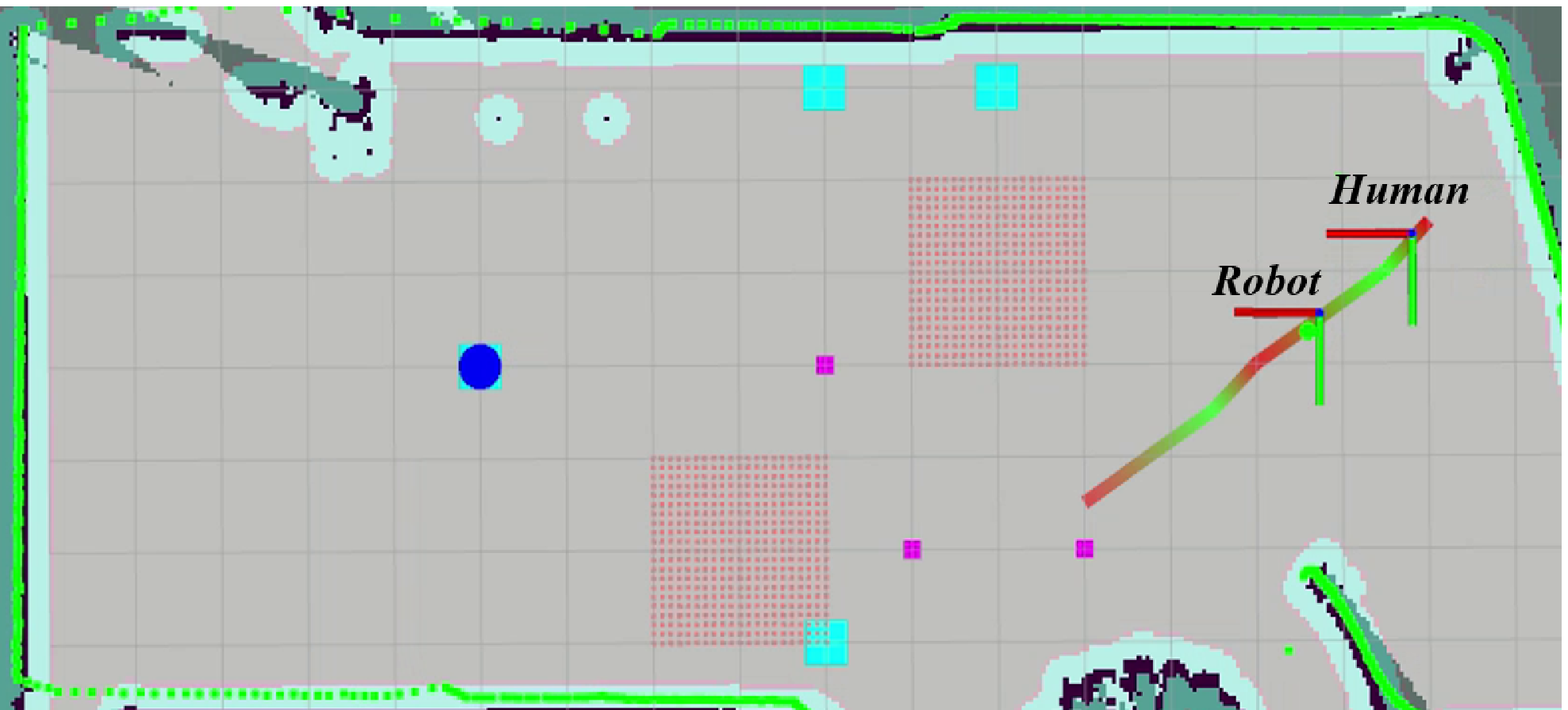}
         \caption{Scene 2}
       \end{subfigure}
       \begin{subfigure}[b]{0.9\linewidth}
         \includegraphics[width=\linewidth]{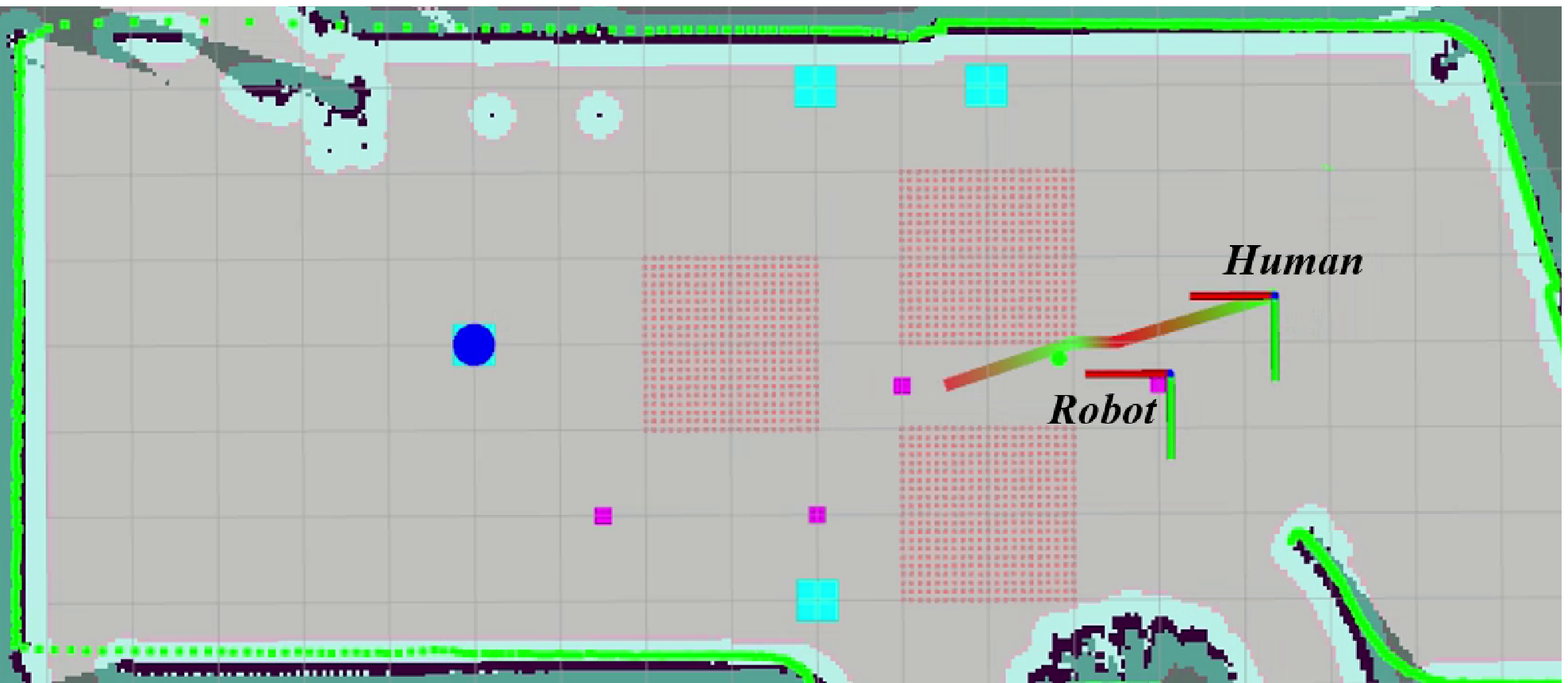}
         \caption{Scene 3}
       \end{subfigure}
      \caption{Guide planning process in three simulated scenes. The space configurations for three scenes are showed in (a), (b) and (c) respectively. Lines filled with red and blue represent the prediction of human future motion and the robot behavior along the way. Red stands for pointing guide and green stands for leading guide. The more distant it is from the current moment, the lighter the color displays. As shown in the figures, the robot would switch its guiding behaviors based on the prediction of human motion. The optimal behavior sequence is produced to better assist the user to avoid context confusing areas in an efficient yet legible way.}
      \label{figure5}
\end{figure}

We ran $50$ independent trials for each scene using our method and the artificial potential field (APF) \cite{IEEEexample:6}, i.e., $150$ trials in total for each method. The result is shown in Table I, Table II, and Table III.
\begin{table}[b!]
\caption{Overall Guide Success Rate in Simulation}
\label{table1}
\begin{center}
\begin{tabular}{cccc}
\toprule
\ & \multirow{2}{*}{\textbf{Scene 1}} & \multirow{2}{*}{\textbf{Scene 2}} & \multirow{2}{*}{\textbf{Scene 3}}\\
\\
\midrule
Our approach & 1.0 & 1.0 & 1.0\\
APF & 1.0 & 0.9 & 1.0\\
\bottomrule
\end{tabular}
\end{center}
\end{table}
\begin{table}[b!]
\caption{Context-Ambiguity Disturbed Ratio in Simulation}
\label{table2}
\begin{center}
\begin{tabular}{cccc}
\toprule
\ & \multirow{2}{*}{\textbf{Scene 1}} & \multirow{2}{*}{\textbf{Scene 2}} & \multirow{2}{*}{\textbf{Scene 3}}\\
\\
\midrule
Our approach & 0.0 & 0.0 & 0.037\\
APF & 0.038 & 0.034 & 0.074\\
\bottomrule
\end{tabular}
\end{center}
\end{table}
\begin{table}[b!]
\caption{Socially Discomfort Time Ratio in Simulation}
\label{table3}
\begin{center}
\begin{tabular}{ccccccc}
\toprule
& \multicolumn{2}{c}{Scene 1} & \multicolumn{2}{c}{Scene 2} & \multicolumn{2}{c}{Scene 3} \\
&  $t_{d_i}$ & $t_{d_p}$ & $t_{d_i}$ & $t_{d_p}$ & $t_{d_i}$ & $t_{d_p}$\\
\midrule
Our approach & 0.0 & 0.076 & 0.0 & 0.070 & 0.0 & 0.062\\
APF & 0.0 & 0.0 & 0.0 & 0.0 & 0.045 & 0.04\\
\bottomrule
\end{tabular}
\end{center}
\end{table}

For the overall guide success rate, our approach managed to guide the human agent to arrive at the destination within time limit in all the trials under every scene set. By comparison, the artificial potential field approach \cite{IEEEexample:6}, although with considerably high success rate, failed in several cases where the deadlock problem happened and exceeded the time limit.

As for the context-ambiguity disturbed time ratio, our approach performed quite well in the first two scenes with no ambiguity disturbed time. When the affordance space configuration becomes more complicated, as setup in scene 3, cases increased where the human agent walked across the inconspicuous affordance spaces. Despite this, the time ratio, valued as $0.037$, was still critically low compared with that when using artificial potential field \cite{IEEEexample:6}, proving better ambiguity distinguishment performance of our approach.

In terms of social comfort, our approach maintained relatively high and stable interaction distance for proper proxemics etiquette. Specifically, the time ratio for the moments below personal distance settled around $0.1$, showed
as $0.076$, $0.07$ and $0.062$ for three scenes respectively. Although it is higher than the result when using artificial potential field \cite{IEEEexample:6}, the ratio itself is at relatively small value. While there exit some time instances when the human-robot distance fell beneath intimate distance via the method of artificial potential field, no case was observed for our approach. This indicates the stable social comfort maintained by our approach.

\subsection{Real-World Experiments}
\begin{figure}[t!]
      \centering
      \includegraphics[width=2.5in]{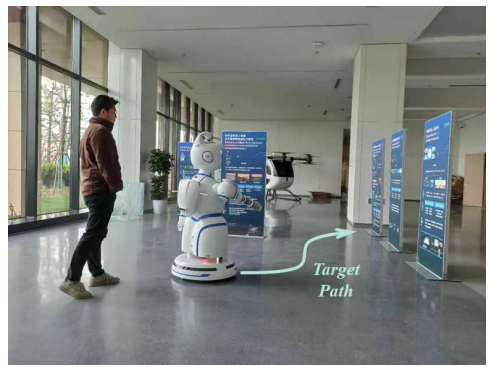}
      \caption{Humanoid mobile robot used in our real-world experiments. The robot is performing pointing guide to help user select the proper path marked in green.}
      \label{figure6}
\end{figure}

\begin{table}[b!]
\caption{Real-World Overall Guide Success Rate}
\label{table4}
\begin{center}
\begin{tabular}{cccc}
\toprule
\ & \multirow{2}{*}{\textbf{Scene 1}} & \multirow{2}{*}{\textbf{Scene 2}} & \multirow{2}{*}{\textbf{Scene 3}}\\
\\
\midrule
Our approach & 1.0 & 1.0 & 1.0\\
APF & 1.0 & 0.92 & 1.0\\
\bottomrule
\end{tabular}
\end{center}
\end{table}
\begin{table}[b!]
\caption{Real-World Context-Ambiguity Disturbed Ratio}
\label{table5}
\begin{center}
\begin{tabular}{cccc}
\toprule
\ & \multirow{2}{*}{\textbf{Scene 1}} & \multirow{2}{*}{\textbf{Scene 2}} & \multirow{2}{*}{\textbf{Scene 3}}\\
\\
\midrule
Our approach & 0.0 & 0.0 & 0.046\\
APF & 0.027 & 0.0060 & 0.16\\
\bottomrule
\end{tabular}
\end{center}
\end{table}
\begin{table}[b!]
\caption{Real-World Socially Discomfort Time Ratio}
\label{table6}
\begin{center}
\begin{tabular}{ccccccc}
\toprule
& \multicolumn{2}{c}{Scene 1} & \multicolumn{2}{c}{Scene 2} & \multicolumn{2}{c}{Scene 3} \\
&  $t_{d_i}$ & $t_{d_p}$ & $t_{d_i}$ & $t_{d_p}$ & $t_{d_i}$ & $t_{d_p}$\\
\midrule
Our approach & 0.0040 & 0.21 & 0.0086 & 0.18 & 0.020 & 0.22\\
APF & 0.011 & 0.25 & 0.013 & 0.26 & 0.017 & 0.22\\
\bottomrule
\end{tabular}
\end{center}
\end{table}

To further validate our approach, real-world experiments were conducted in the same exhibition hall with same obstacle settings as our simulation experiments did. We carried out the experiments using our customized human-like guide robot ZJRobot which is equipped a mobile base, two arms of 6 DoF, and speakers (Fig. 6). A UWB-based localization system was deployed to track human positions in real-time.

We invited $14$ human subjects for independent experiments and each of them went through tour-guides under three different scene sets with different affordance space topologies in a counterbalance order. The result shows are shown in Table IV, Table V and Table VI.

Similar to that in the simulation, our approach succeeded in all the trials, better than the results of artificial potential field where $1$ time-out failure happened in scene 2. For the performance in distinguishing context ambiguity, our approach also outperformed the counterpart in all three scenes, with ambiguity disturbed time ratio of $0.0$, $0.0$ and $0.046$ respectively, which are significantly lower than that when the guide was supported by the artificial potential field.

When it comes to social comfort, our approach conformed to the social compliance through most of the time, with the time ratio of $0.21$, $0.18$ and $0.22$ when the distance is below the personal distance and valued as $0.0040$, $0.0086$ and $0.020$ it is below the intimate distance, better than the social-compliance performance using artificial potential field. The improved functionality of our approach to proactively distinguish context ambiguity, stably maintain social comfort and timely guide humans toward the destination was therefore proved in both simulation and reality.

\section{Conclusion}

In this paper, we present a multi-behavior guide planning framework for providing proactive guidance. By considering the influence of robot behaviors, our method can predict human motion more accurately, which allows our behaviour planner to achieve balance between efficiency and legibility. The proposed framework was tested in both simulation and real-world experiments. Compared with the artificial potential field method, our work achieved higher success rate and less context-ambiguity disturbed time. Furthermore, stable social comfort could also be maintained across various scenarios of ambiguous contexts. There still remains some space for further improvement. For example, taking account more types of guiding behaviors into our framework can bring better user experience in tour guide.



\addtolength{\textheight}{-12cm}   





\bibliographystyle{IEEEtran}
\bibliography{IEEEabrv,IEEEexample}

\end{document}